\documentclass[review]{elsarticle}
\usepackage{lineno}
\usepackage{hyperref}
\modulolinenumbers[5]

\usepackage{graphicx}
\usepackage{multirow}
\usepackage{subfigure}

\journal{Journal of \LaTeX\ Templates}

\begin{document}

\begin{frontmatter}

\title{Feature Learning for Accelerometer based Gait Recognition}


\author[1]{Szil\'{a}rd Nemes}
\author[2]{Margit Antal} 

\address[1]{Faculty of Mathematics and Computer Science, Babes–Bolyai University, Cluj-Napoca, Romania}
\address[2]{Sapientia Hungarian University of Transylvania, Tirgu Mures, Soseaua Sighisoarei 1C, Tirgu Mures and 540485, Romania}





\begin{abstract}
Recent advances in pattern matching, such as speech or object recognition support the viability of feature learning with deep learning solutions for gait recognition. Past papers have evaluated deep neural networks trained in a supervised manner for this task. In this work, we investigated both supervised and unsupervised approaches. Feature extractors using similar architectures incorporated into end-to-end models and autoencoders were compared based on their ability of learning good representations for a gait verification system. Both feature extractors were trained on the IDNet dataset then used for feature extraction on the ZJU-GaitAccel dataset. Results show that autoencoders are very close to discriminative end-to-end models with regards to their feature learning ability and that fully convolutional models are able to learn good feature representations, regardless of the training strategy.

\end{abstract}


\end{frontmatter}


\section{Introduction}

Murray \cite{MURRAY1967} was the first to explore the potential of gait for distinguishing individuals. She discovered several features that were consistent for one individual but varied between subjects. Subsequent research showed that humans are able to recognize others by their gait \cite{CUTTING1977}. Due to its reliability, gait has been used both as tool for criminal investigation \cite{LARSEN2006} and forensic evidence in court \cite{CONNOR2018}.

Gafurov et al. \cite{GAFUROV2007} identified three modalities of biometric gait recognition: machine vision, floor sensor and wearable sensor-based. Some studies also explored acoustic features for this purpose. While most of the research has been done in the field of machine vision, wearable sensor-based recognition is gaining relevancy due to the availability of built-in inertial sensors in smartphones and watches.

Machine learning-based recognition solutions work with features extracted from the gait data. Many studies use handcrafted features for evaluation, but there are some that already employ neural networks for this task \cite{GADALETA2018} \cite{GIORGI2018}. Connor and Ross \cite{CONNOR2018} state that ``perhaps, rather than collecting a set of handcrafted features from a multimodal gait dataset, a deep learning approach may discover a more effective combination of multimodal features or provide a new and more discriminating view on gait data''.

One of the most important characteristics of deep learning algorithms is their ability to exploit the unknown structure in the input distribution to discover good representations, often at several levels. The discovery of representations, which is often named also feature learning, can be controlled in several ways \cite{BENGIO2011}. There are two main approaches for feature learning: unsupervised and supervised. Given an unlabeled dataset with no class information provided, only unsupervised learning can be applied. However, with a labeled dataset both options become available.

In this paper, we examine these two types of feature learning methods for a gait-based biometric verification system. Automatic feature learning is always present in neural networks trained for classification purposes. These end-to-end models trained in a supervised manner are able to conduct the feature learning in such a way as to discriminate between inputs from different classes. In contrast to these end-to-end models, autoencoders are neural networks designed to learn efficient data encoding (representation) in an unsupervised manner by minimizing the reconstruction error. To the best of our knowledge, this is the first paper investigating unsupervised feature learning for gait recognition; most of the literature investigates automatic feature extraction using supervised learning \cite{GIORGI2017, GIORGI2018, GADALETA2018}.

The main contributions of this paper are as follows: (i) we designed a gait verification system for comparing different types of features extracted from gait signal; (ii) we compared the performance metrics of systems using unsupervised feature learning (autoencoders) to supervised feature learning (end-to-end models); (iii) we investigated the effect of augmentation over verification performance; (iv) we performed visualisations using t-SNE plots to explain the discriminative properties of the learned features. 

The rest of the paper is structured as follows. In Section 2, we present a short literature review highlighting the relevant studies. Section 3 presents how preprocessing methods were applied on the gait signal, and section 4 explains our feature learning approaches. The two proposed gait verification system architectures are presented in section 5. In section 6 the two proposed architectures are evaluated and compared using two benchmark datasets. Finally, the last section concludes the paper.

\section{Literature review}

One of the early works in the field of gait recognition with wearable sensors was done by the VTT Technical Research Center of Finland \cite{AILISTO2005}. They employed a template matching-based solution by computing cross-correlation between extracted cycles and the stored template and comparing the output similarity score with a set threshold for decision making. While they recorded signals on all three axes, only forward-backward and vertical motions were used of the sensor fastened to the waist. An equal error rate (EER) of 6.4\% was achieved for 36 test subjects.

Gafurov et al. \cite{GAFUROV2006} used a sensor positioned in the lower leg's area and obtained EERs of 5\% and 9\% for histogram similarity and cycle length methods respectively.

Lu et al. \cite{LU2014} opted for a statistical modeling solution instead of the previous template matching ones. They aimed for providing a solution that is agnostic of sensor placement and capable of running in real-time on an Android phone. The first phase of feature extraction was used as input for detecting walking segments and the second phase (autocorrelation features) was used for training a GMM-UBM (Gaussian Mixture Model - Universal Background Model). Frames of 512 samples measured at 100 Hz were used with 50\% overlap and the data was collected with a variety of devices: Samsung Galaxy S3, S4, LG Nexus 5 and Intel Xolo. They obtained an EER of 14\% after 20\% of the data was used for training; further increasing this rate had little impact on performance.


Several studies investigated convolutional neural networks trained as end-to-end models for gait classification. Nguyen et al. \cite{NGUYEN2017} used a multi-region size CNN to recogonize users by their gait patterns using accelerometer and gyroscope sensors. Their method resulted in a 10.43\% EER on the OU-ISIR dataset \cite{OUISIR}. Zhao and Zhou \cite{ZHAO2017} proposed an image-based gait recognition model using a CNN model, which was trained by images obtained from gait data sequences. Dehzangi et al. \cite{DEHZANGI2017} proposed a novel approach for human gait identification using time-frequency expansion of human gait cycles. They trained a 2D CNN and evaluated gait identification performance on a private dataset containing gait recordings from 10 subjects, and obtained 97.06\% subject identification accuracy. Delgado et al. \cite{DELGADO2019} proposed another end-to-end approach based on CNN architectures for gait authentication and recognition that uses raw inertial data as input. Comparing the performance of systems using only the acceleration sensor and those using only the gyroscope, it was found that the acceleration sensor-based system is more accurate. Furthermore, the system using both sensors performed better in all cases.

The first attempt to use deep learning for feature extraction was the work of Gadaleta and Rossi \cite{GADALETA2018}. They obtained an EER of only 0.15\% and report that gait recognition can be reliably performed considering five consecutive step cycles. A CNN (convolutional neural network) was trained once on a representative set of users using both accelerometer and gyroscope data and used to extract features from further step cycles. A one-class SVM was used for gait recognition. Their IDNet dataset of 50 subjects is publicly available\footnote{\url{http://signet.dei.unipd.it/human-sensing/}}.

Other types of deep learning models were also proposed for gait recognition. Gao et al. \cite{GAO2019} proposed a combined long short-term memory (LSTM) and CNN for abnormal gait classification and reported 93.1\% accuracy. More recently, Fernandez et al. \cite{FERNANDEZ2019} proposed  Recurrent Neural Network (RNN) to generate feature vectors. Their gait verification system resulted in 7.55\% EER on the OU-ISIR dataset using a template-based model.

Zhang et al. \cite{ZHANG2015} presented the ZJU-GaitAcc dataset with only accelerometer data at five body locations and measured a 95.8\% accuracy for identification and 2.2\% EER for verification when considering input from all sensors. Marsico and Mecca \cite{MARSICO2017} measured on both this dataset and OU-ISIR and made use of a template matching solution with DTW (dynamic time warping). They obtained a 93\% recognition rate for identification when comparing entire walks to templates that contained a similar number of steps and 83\% when comparing ``best steps'' (centroid of current step cycles cluster) to stored templates.

The usage of deep learning for this dataset was proposed by Giorgi et al. \cite{GIORGI2017}. A CNN was used for feature extraction and classification taking as input accelerometer data of all three channels from all five sensors. A rank 1 accuracy (the first correct label has the highest probability) of 92\% is reported for a closed set of 153 subjects within one session (same-day). This idea is improved in \cite{GIORGI2018} by complementing the CNN architecture with a GRU (gated recurrent unit) for feature extraction and enabling it to evaluate features of individual sensors. All five sensors considered, a same-day identification accuracy of 99.06\% was reported in the same-day scenario and 88\% considering only the sensor on the right side of the pelvis. Cross-day in this work denotes that the model was trained with data from both sessions of the dataset; these scores are slightly lower than in the same-day setting.

In a previous work \cite{NEMES2018}, we investigated several questions in gait recognition on ZJU-GaitAcc dataset using statistical features. We concluded that at least five consecutive steps are reliable for authentication (1.58\% EER) and that impostor selection for testing has a significant impact on the evaluation.


In a recent work Tran and Choi \cite{TRAN2020} investigated the effect of data augmentation on training deep neural networks for gait recognition. They concluded that applying data augmentation such as arbitrary time deformation or stochastic magnitude perturbation for creating more training data, increased the recognition performance effectively.

While the usage of deep learning is promising in the field of gait recognition, autoencoders so far have not been used for feature extraction in this context. This work explores their viability for this task by comparing autoencoder-extracted features with raw data and end-to-end model-extracted features for evaluation.

\section{Signal processing}

It is recommended to segment gait signal to obtain enough samples for modeling gait data. Sprager and Juric \cite{SPRAGER2015} divided gait recognition approaches based on segmentation type into frame-based and cycle-based methods. One gait cycle is defined as the time interval starting when one foot makes initial contact with the ground and ending when the same foot touches the ground again \cite{SPRAGER2015}. In contrast with cycle-based approaches, frame-based approaches work with short, fixed-length frames. In addition, these can be either overlapping or non-overlapping. The frame length is usually chosen to cover a complete gait cycle.

In this section, we cover three key questions: (i) why frames were chosen over cycles; (ii) how the frame length was determined; (iii) how we brought the two datasets to the same sampling frequency.

Firstly, with regards to the question of working with a cycle-based or frame-based system, this was mainly decided by the datasets involved in this work: ZJU-GaitAcc \cite{ZHANG2015} and IDNet \cite{GADALETA2018}. The former also provides manual annotations of cycle edges, while the latter provides no endpoints for step cycles. Since IDNet is not an annotated dataset and segmentation for the step cycle would have brought errors into the system \cite{DEB2020}, we decided to choose frame-based modeling.

\begin{figure}
\centering
\includegraphics[width=8cm]{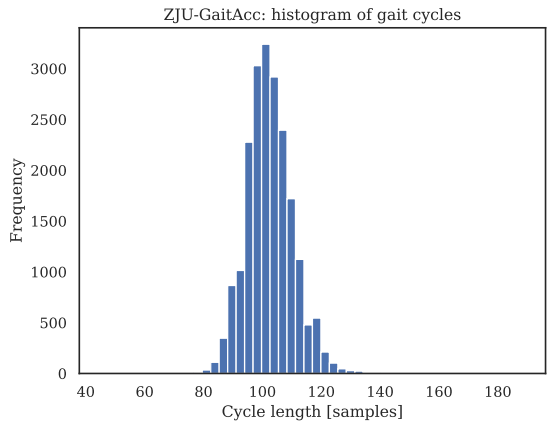}
\caption{Cycle length histogram of the ZJU-GaitAcc dataset.}
\label{fig:cycleshisto}
\end{figure}

Secondly, the length of a frame was to be determined so that it covered complete step cycles. A histogram of step cycles was prepared for the annotated dataset having this goal in mind. Figure \ref{fig:cycleshisto} illustrates this histogram of the whole dataset (average: 102 samples, median: 110 samples; sampling frequency 100 Hz). Considering this, we chose 128 as the frame length (1.28 seconds), which covers 99.5\% of gait cycles of the dataset. In this paper, we work with non-overlapping frames.

Lastly, we solved the non-uniform sampling frequency problem of the IDNet dataset by resampling it uniformly at 100 Hz frequency. As a final step, z-score normalization was applied to each frame.

\section{Feature learning}

In this section, we briefly present how convolutional neural networks and autoencoders learn. Here we also explain why we opted for a Fully Convolutional Neural Network (FCN) as end-to-end model and how this architecture’s equivalent in the unsupervised setting was constructed as an autoencoder.

The performance of a machine learning algorithm can benefit the most from thorough feature engineering. This task, however, requires domain expertise, therefore it is highly desirable to design algorithms that are able to extract and organize the discriminative information from the data automatically \cite{BENGIO2013}. Feature learning or representation learning is the process that allows a system to automatically discover data representation in a different dimension than that of raw data. This summarization process, be it supervised or unsupervised, is expected to facilitate better classification performance. There are various representation learning methods, but in this paper we focus on deep learning based methods. As emphasized by Bengio et al. \cite{BENGIO2013} it is not easy to formulate good criteria for representation learning. In case of classification, the objective is obvious: minimize misclassification on the training dataset.

Deep convolutional neural networks (CNN) are extremely popular for image recognition. Applying a convolution over time series is the equivalent of using a sliding filter. These filters are one-dimensional and can be seen as a non-linear transformation of a time series. However, when considering multidimensional times series (e.g. accelerometer data), the filter has an extra dimension equal to the dimensionality of the input signal. Training such a network consists of tuning the parameters of filters applied on input data. Applying several filters on a time series is the equivalent of learning several representative features of the time series. One can also introduce different pooling layers with the goal of downsampling and apply them either locally or globally. Local pooling takes an input time series and reduces its length by aggregating (average or max) over a sliding window of the time series. Global pooling aggregates the time series into a single value.

An autoencoder is a special type of neural network with the purpose of mimicking its input with its output. This is usually done by bringing data to a latent feature space (\emph{encoding}) and trying to reconstruct the input from it (\emph{decoding}). Formally, this is the equivalent of $g(f(x)) = x, f:{R}^{N}\rightarrow{R}^{d}, g: {R}^{d}\rightarrow{R}^{N}$, where $N$ is the dimension of the original feature space, $d$ is the dimension of the latent space and $x\in{{R}^{N}}$. Their traditional role is dimensionality reduction and feature learning \cite{HINTON2006}. The way an autoencoder learns is by minimizing the loss function $L(x, g(f(x)))$, where $L$ denotes the dissimilarity between the input and the output (e.g. Mean Squared Error)
\cite{GOODFELLOW2016}.

\section{Gait verification system}

In this section, we describe two architectures for biometric verification systems. Both of them first go through an automatic feature learning stage. However, while one employs supervised learning for feature learning, the other one is does this unsupervised.

\begin{figure*}
\centering
\includegraphics[width=1\textwidth]{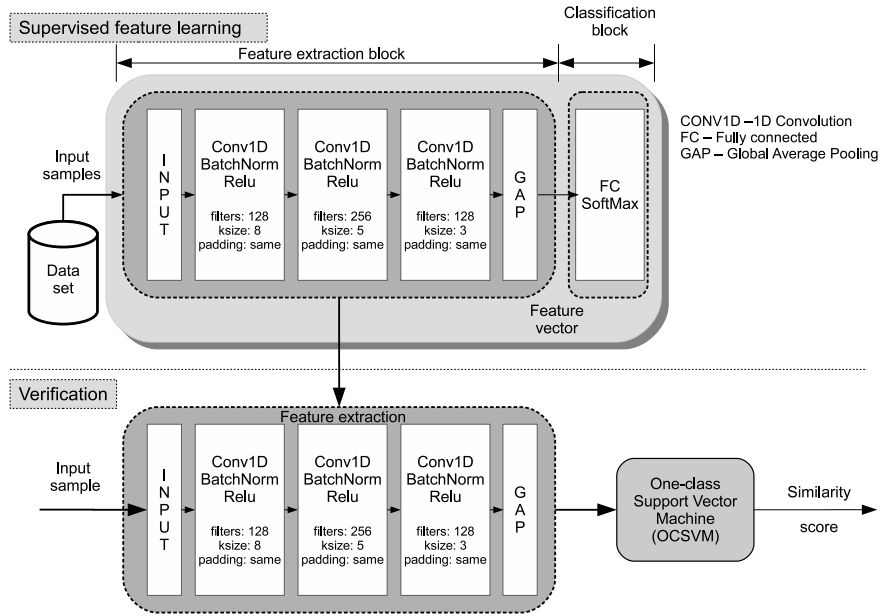}
\caption{Architecture of our biometric verification system using
supervised feature learning.}
\label{fig:endtoend_pipeline}
\end{figure*}

Figure \ref{fig:endtoend_pipeline} shows the architecture of our system based on supervised feature learning. In the first stage, an FCN model is trained using a labeled dataset for classification. The Global Average Pooling (GAP) layer outputs the features extracted by the convolutional blocks denoted as $f=(f_1, f_2, \ldots, f_D)$. The last Fully Connected (FC) layer provides an output vector with size equal to the number of classes in the dataset $y =(y_1, y_2,\ldots, y_K)$. After the network has been trained, the last layer is removed and the resulting network can be used as a universal feature extractor.

In the second stage, a one-class Support Vector Machine (OCSVM) is trained for each user by using the feature vectors extracted from the positive samples of that given user. This model will return a similarity score for each testing sample based on which the system will accept or reject the sample. Using OCSVM for gait verification was proposed by Gadaleta and Rossi \cite{GADALETA2018}, then used in a recent work\cite{TRAN2020}.

\begin{figure*}
\centering
\includegraphics[width=1\textwidth]{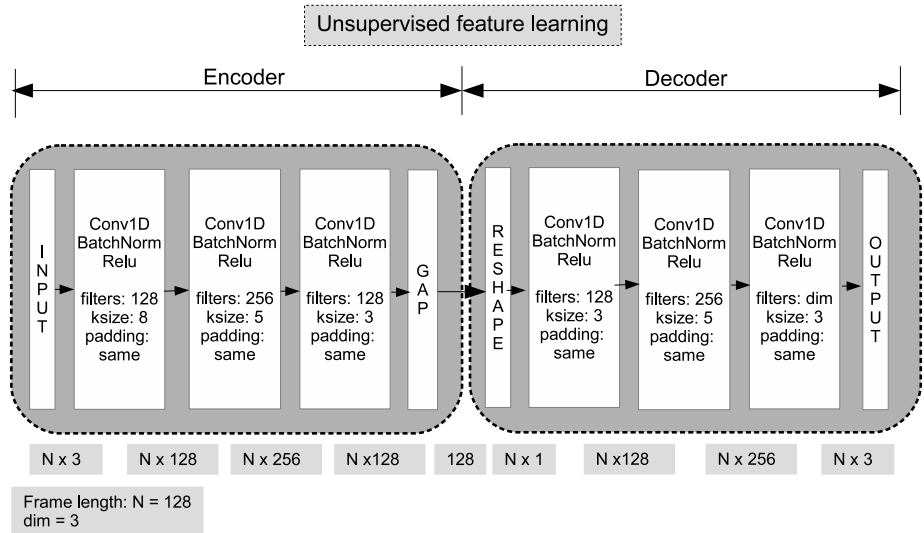}
\caption{Architecture of the convolutional autoencoder.}
\label{fig:autoencoder_pipeline}
\end{figure*}

The architecture of our autoencoder used for unsupervised feature learning is depicted in Figure \ref{fig:autoencoder_pipeline}. 

The autoencoder consists of an encoder and a decoder. For the encoder part, the same FCN architecture was used as in the supervised setting, except for the last fully connected layer. Class information was removed from a designated dataset and used for training. Following this, the encoder part was saved and used later for feature extraction. The verification part is the same as for the supervised feature learning shown in the bottom part of Figure \ref{fig:endtoend_pipeline}.


In this work, we designed automatic feature extractors for a gait recognition system. These feature extractors are built in two ways: (i) training an autoencoder without using the class information of the training data - this is the unsupervised or generative way, (ii) training a deep neural model with the usage of class information of the training data - the supervised or discriminative way. Usually, a deep neural network trained for classification consists of several layers responsible for feature learning and its last layer tasked with classification. In this case, the features learned by the neural network will be discriminative features for the given classification task, as the weights are tuned according to a loss function (e.g. categorical crossentropy) that penalizes classification error.

Fawaz et al. \cite{FAWAZ2019} implemented nine end-to-end models in their open source deep learning framework and evaluated them on several multidimensional time series datasets. The FCN model performed best, followed by the ResNet model \cite{WANG2017}. Considering these results, the FCN model was chosen also for our work. FCN models contain pure convolutional layers without local pooling layers, therefore the length of the time series does not change between convolutions. The original architecture proposed by Wang \cite{WANG2017} is composed of three convolutional blocks. Each convolutional block contains a convolution, a batch normalization and ReLu activation. The first block contains 128 filters (kernel size: 8), the second block 256 filters (kernel size: 5), and the last block 128 filters (kernel size: 3). After the third block, global average pooling is applied. All convolutions preserve the length of the time series (stride: 1, zero padding). This construction is the core of the end-to-end model, having only another fully connected layer with an additional softmax applied on its output for obtaining class probabilities. It is also the foundation of the autoencoder architecture: the encoder part is the same as the FCN model, while the decoder part consists of the same layers, but in reverse order.

The IDNet dataset was used for feature learning using a train-validation split of 60\%-40\%. The models (both end-to-end and autoencoder) were trained in Keras \cite{KERAS} for 100 epochs having a model checkpoint for saving the model with the lowest validation loss. All models were optimized using the Adam optimizer, which is a variant of Stochastic Gradient Descent (SGD) \cite{KINGMA2015}. The learning rate was reduced by a factor of 0.5 when the model's training loss has not improved in 50 consecutive epochs (with a minimum value equal to 0.0001) \cite{FAWAZ2019}. Having finished training, the last layer of the end-to-end FCN model was removed and saved as such with the aim of using it later for feature extraction. In case of the trained autoencoder, the decoding part was removed, leaving the encoder with the task of feature extraction.

\section{Experimental study}

In this section, we describe our experiments designed to compare the two types of feature learning for the same biometric gait verification system. We compare the efficiency of the two feature learning methods by equal error rates (EER) and area under the curve (AUC) in order to be comparable to previous work.

\subsection{Datasets}

The ZJU-GaitAcc dataset \cite{ZHANG2015} contains gait data recorded from 175 subjects. Two thirds of the subjects are male, subject height is in the range of 1.5 to 1.9 meters with ages from 16 to 40. The walking of the subjects was recorded by 5 accelerometers fastened to different locations on the body: right wrist, left upper arm, right side of pelvis, left thigh and right ankle. In this work, the only sensor considered is the one fastened to the right side of the pelvis. We decided to use only this sensor in order to resemble the real world scenario, namely using smartphones to capture gait data (people usually carry theirs in their right pocket).

The dataset is split into three sessions. Session 0 contains data from 22 subjects. The walking data of these subjects were recorded in a single session. The remaining 153 subjects participated in two sessions, these are session 1 and session 2. Each subject has 6 recordings in a given session, resampled at 100 Hz. The recordings were manually annotated (the start and end positions of the step cycles are marked). Each recording contains 7-14 full step cycles with a usual cycle containing 90-118 samples. The dataset contains 6.59 hours of walking data (session 0: 29.90 minutes, session 1: 185.84 minutes, session 2: 181.96 minutes).

The IDNet dataset \cite{GADALETA2018} was collected from 50 subjects during a period of six months. The subjects wore an Android smartphone in their right front pocket. Six types of Android devices were used for data collection. 17 subjects recorded their walking data in a single session, while the number of sessions recorded by the remaining 33 subjects was between 2 and 14. The sessions are of variable lengths with an average of 6.77 minutes with a standard deviation of 2.41. The dataset contains 15.23 hours of walking data.

The most important properties of the two datasets are summarized in Table \ref{tab:gaitdatasets}. While the ZJUGaitAccel dataset was collected by Wii Remote devices and the final dataset was resampled at exactly 100 Hz, the IDNet dataset contains data collected by a variety of mobile phones. The sampling frequency for this dataset varies between 43.69 and 198.59 Hz, on average 122.60 Hz. We resampled this dataset at 100 Hz using linear interpolation to match the frequency of the other dataset.

\begin{table}
\centering
\caption{Basic characteristics of the ZJU-GaitAcc and IDNet datasets}
\label{tab:gaitdatasets}       
\begin{tabular}{lllll}
\hline
Dataset & Frequency (Hz) & Subjects & Total     & Subject's average\\
         & average(stdev) &             & time (h) & time (min) \\
\hline
ZJU-GaitAcc & 100 (0)       & 175       & 6.59  & 2.22 (0.19)\\
IDNet       & 122.60 (30.70) & 50       & 15.23 & 18.28 (21.14)\\
\hline
\end{tabular}
\end{table}

\subsection{Measurement protocol}

We decided which dataset to use for feature learning and which to use for evaluation by considering two aspects. The first aspect was the amount of data available, the second aspect was the variety of devices used in data collection. Therefore, the IDNet dataset is used for the feature learning phase, as it has roughly three times more data recorded with multiple devices, hence the model can generalize better, while the ZJU-GaiAcc dataset is used for user verification evaluations.

Two protocols were defined for user verification measurements. Firstly, we have the same-day measurements, when both the training and test data are taken from the same session. We report the results separately for session 1 and 2, each containing data from the same 153 subjects. Secondly, we have the cross-day measurements, when the training data is taken from session 1 and the testing data is taken from session 2.

As a first step, we trained the neural networks performing the feature extraction using the IDNet dataset. As a second step, we extracted the features from the frames of the ZJU-GaitAcc dataset with the trained feature extraction networks. In the third step, we trained an OCSVM model for each subject of the ZJU-GaitAcc dataset by using two-thirds of the available data to create the models. In the final step, we tested the models: each subject model was tested using the remaining one-third of the data as positive test data and all available data from all other subjects as negative data. This is how the measurement was performed for the same-day evaluation protocol. In case of cross-day measurements, the OCSVM model was trained using the data of a subject from session 1, while for testing we used both the positive and negative data from session 2.

\subsection{Results}

The results presented in this section can be reproduced using our public GitHub repository \footnote{\url{https://github.com/margitantal68/featurelearning}}.

\subsubsection{Baseline results}

The first experiment was to compare our proposed feature extraction approaches to using directly raw data as features. Both the supervised and unsupervised setups extract features from a single frame of gait data, which consists of 128 x 3 = 384 raw samples. When applying them, the input shape of the data is preserved and provide an output vector of 128 features. When we work directly with raw data, the three directional acceleration vectors are concatenated, resulting in a vector of 384 features. Table \ref{tab:verification} shows the performance comparison summary. AUC values show that both types of feature learning improved the system performance by 31\% and 34\% in case of same-day evaluation, and by 14\% and 24\% in cross-day settings. A similar magnitude improvement can be observed along the EER metrics. In this case, the system with the smaller error is considered better. Along both metrics, the system working with supervised feature learning performed better.

\begin{table}
\centering
\caption{Performances of verification systems using different types of features. Features: AE - autoencoder, EE - end-to-end; Evaluation:  SD-S1 - same-day session1, SD-S2 - same-day session2, CD - cross-day.}
\label{tab:verification}       
\begin{tabular}{cccc}
\hline
Features&	Evaluation&	Avg AUC[\%]& Avg EER[\%]\\
\hline

\multirow{3}{*}{raw}& SD-S1&	60.63 (23.36)&	41.33 (20.24)\\
					& SD-S2&	63.37 (24.36)&	38.66 (21.32)\\
					& CD   &	54.38 (26.01)&	46.64 (22.51)\\
\hline					

\multirow{3}{*}{AE} & SD-S1&	91.30 (10.21)&	14.75 (10.84)\\
					& SD-S2&	93.92 (8.46)&	11.77 (9.70)\\
					& CD   &	68.56 (25.11)&	35.99 (21.58)\\
\hline
\multirow{3}{*}{EE} & SD-S1&	95.27 (7.66)&	9.32 (9.02)\\
					& SD-S2&	97.22 (4.97)&	6.96 (7.29)\\
					& CD   &	78.56 (21.90)&	26.58 (19.53)\\
\hline

\end{tabular}
\end{table}

\begin{figure*}
\centering
\begin{subfigure}
\centering
\includegraphics[width=0.48\textwidth]{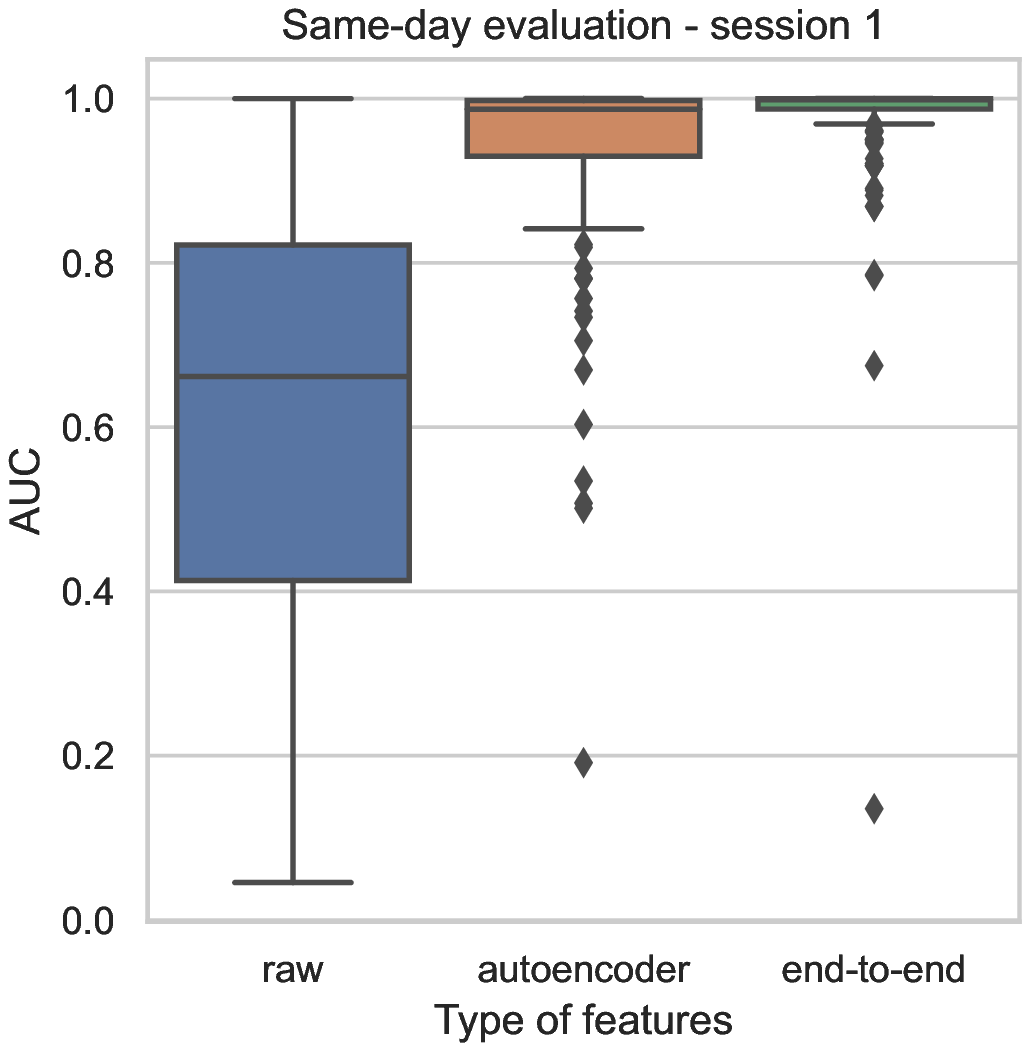}
\end{subfigure}
\begin{subfigure}
\centering
\includegraphics[width=0.48\textwidth]{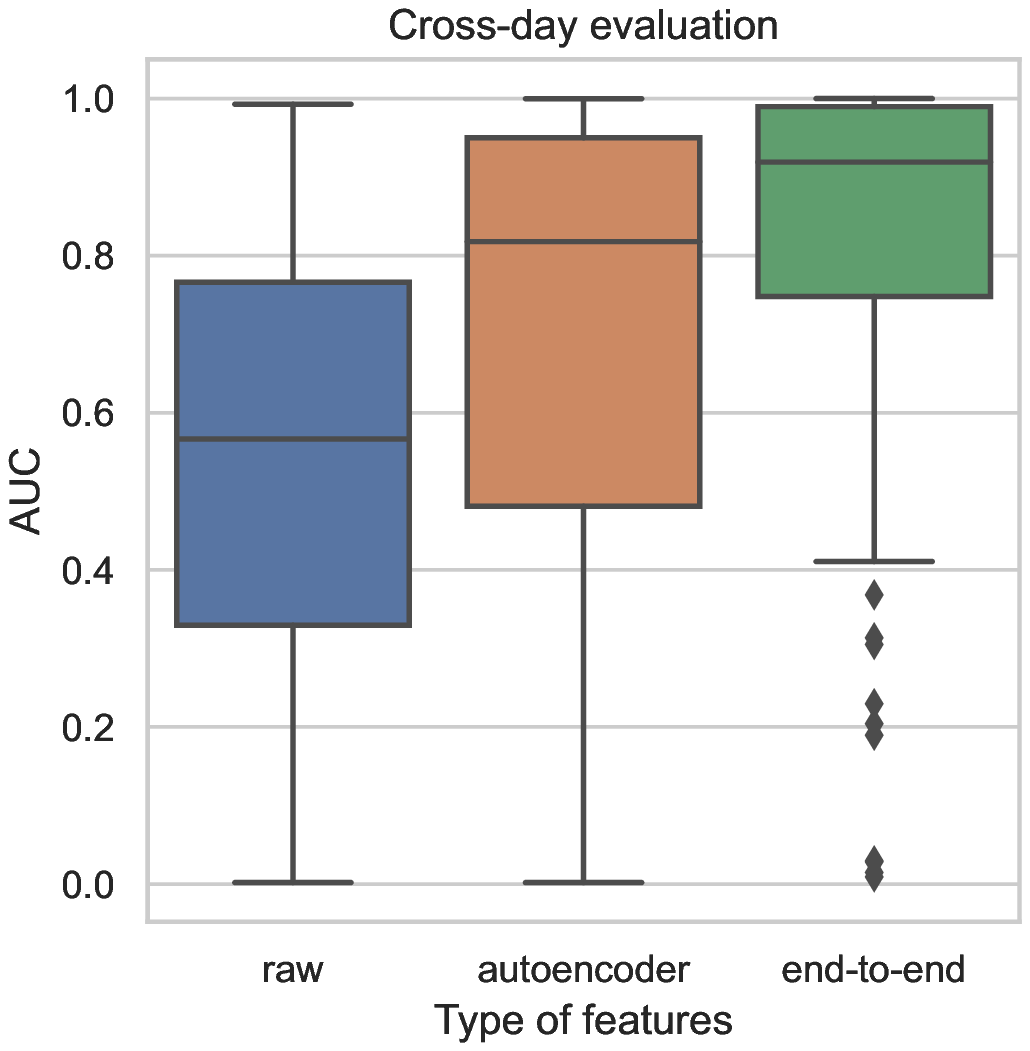}
\caption{Same-day vs. cross-day evaluation results for the three types
of features: raw, autoencoder, end-to-end.}
\label{fig:evaluation}
\end{subfigure}
\end{figure*}

Figure \ref{fig:evaluation} shows the distribution of users’ AUC values obtained in same-day, repectively in cross-day settings. Here we can see that the system based on supervised learning has not only the best average performance, but also the least performance variance among subjects. It is also important to note that for cross-day measurements system performances are poor for many subjects.

\subsubsection{The effect of augmentation on feature learning}

Data augmentation is a technique used to improve the model's generalization ability and to prevent overfitting \cite{GOODFELLOW2016}. In this paper, we used two types of augmentation. The first type of augmentation proposed by Giorgi et al. \cite{GIORGI2018} adds a random noise in the range $[-0.2, 0.2]$ to each dimension of the three-dimensional signal. We propose another type of augmentation that fits frame-based segmentation. Let us consider a one-dimensional signal $X={\{x_1, x_2, \ldots, x_n\}}$. A random position $k$ is generated in the interval of $2..n-1$, then the signal is shifted circularly to the left resulting in ${\{x_k, x_{k+1}, \ldots,x_{n}, x_1, \ldots, x_{k-1}\}}$. The same circular shifting is applied to each dimension. Augmentation was applied to each frame, doubling the size of the training dataset.

Table \ref{tab:augmentation} summarizes the performances obtained for using data augmentation in the feature learning stage. Both types of augmentations had only a limited impact on the performance of the autoencoder-based feature extractor. By applying the first type of augmentation, the performance of this system increased by 1.5\% in terms of AUC, but only in the cross-day setting. Furthermore, the second type of augmentation improved the system's performance both in same-day and cross-day settings. In contrast, augmentation had no significant effect on the supervised feature extractor. In the same-day settings, performance did not change, while in the cross-day settings it even decreased by 2\%.

\begin{table*}
\centering
\caption{The effect on augmentation over feature learning. Features: AE - autoencoder, EE - end-to-end; Evaluation:  SD-S1 - same-day session1, SD-S2 - same-day session2, CD - cross-day; Augmentation: cshift - circular shift, rnd - random noise. }
\label{tab:augmentation}
\begin{tabular}{ccccc}
\hline
Features&			Augmentation& 				Evaluation&	Avg AUC[\%]& Avg EER[\%]\\
\hline

\multirow{6}{*}{AE} & \multirow{3}{*}{rnd}   & SD-S1&	91.12 (9.71)&	14.46 (10.12)\\
					& 					     & SD-S2&	93.74 (8.17)&	11.98 (9.25)\\
					& 					     & CD   &	69.91 (24.73)&	34.71 (21.38)\\
\cline{2-5}
					& \multirow{3}{*}{cshift}& SD-S1&	93.94 (7.61)&	11.04 (8.69)\\
					& 					  	 & SD-S2&	95.83 (6.58)&	8.70 (7.88)\\
					& 					  	 & CD   &	74.30 (24.32)&	30.43 (21.54\\
\hline
\multirow{6}{*}{EE} & \multirow{3}{*}{rnd}   & SD-S1&	95.24 (7.92)&	9.26 (9.18)\\
					& 					     & SD-S2&	97.14 (4.90)&	6.59 (6.19)\\
					& 					     & CD   &	76.93 (23.51)&	27.61 (20.85)\\
\cline{2-5}
					& \multirow{3}{*}{cshift}& SD-S1&	95.58 (7.69)&	8.54 (8.25)\\
					& 					  	 & SD-S2&	97.52 (4.66)&	6.40 (6.55)\\
					& 					  	 & CD   &	76.87 (23.96)&	27.55 (21.25)\\

\hline

\end{tabular}
\end{table*}

\subsubsection{Visualisations}

Figure {fig:visualisations} show samples distribution from 10 subjects visualized with t-SNE \cite{TSNE} using raw data, autoencoder and end-to-end system features. As seen for raw data, subject samples are not well separated i.e. do not form distinct clusters. Both autoencoders and end-to-end systems produced features that separate well the gait segments of the 10 subjects.

\begin{figure*}
\centering
\begin{subfigure}
\centering
\includegraphics[width=0.32\textwidth]{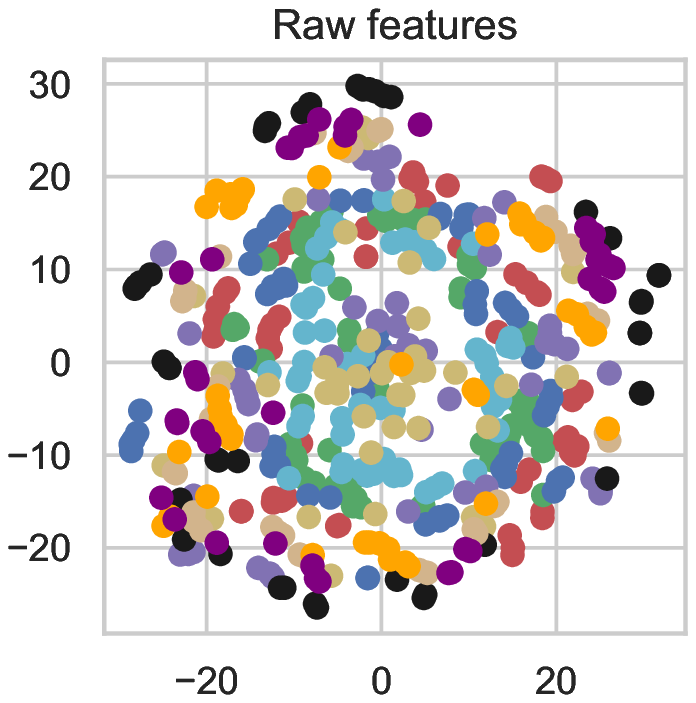}
\end{subfigure}
\begin{subfigure}
\centering
\includegraphics[width=0.32\textwidth]{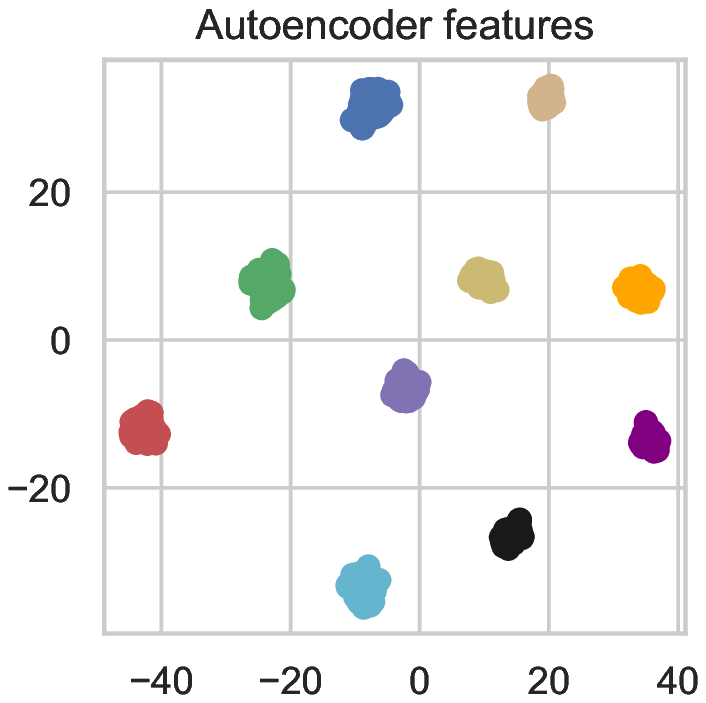}
\end{subfigure}
\begin{subfigure}
\centering
\includegraphics[width=0.32\textwidth]{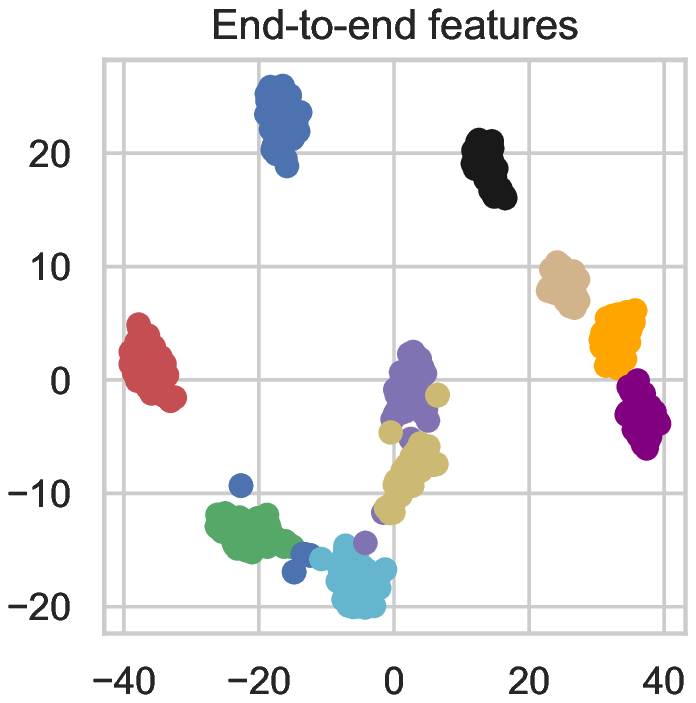}
\end{subfigure}
\caption{The first 10 subjects of session 0 using the three types of
features: raw, autoencoder, end-to-end.}
\label{fig:visualisations}
\end{figure*}

\subsubsection{Score aggregation}

\begin{figure*}
\begin{subfigure}
\centering
\includegraphics[width=0.48\textwidth]{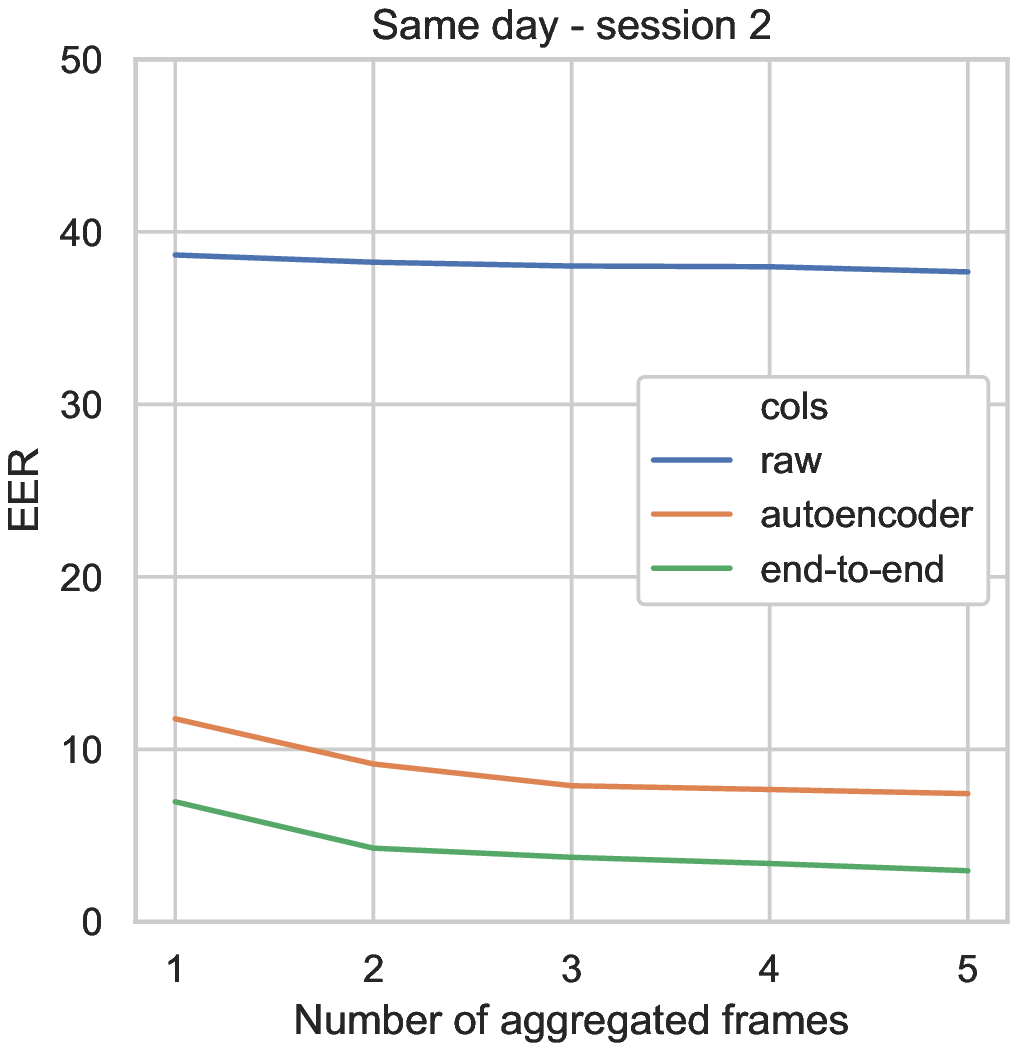}
\end{subfigure}
\begin{subfigure}
\centering
\includegraphics[width=0.48\textwidth]{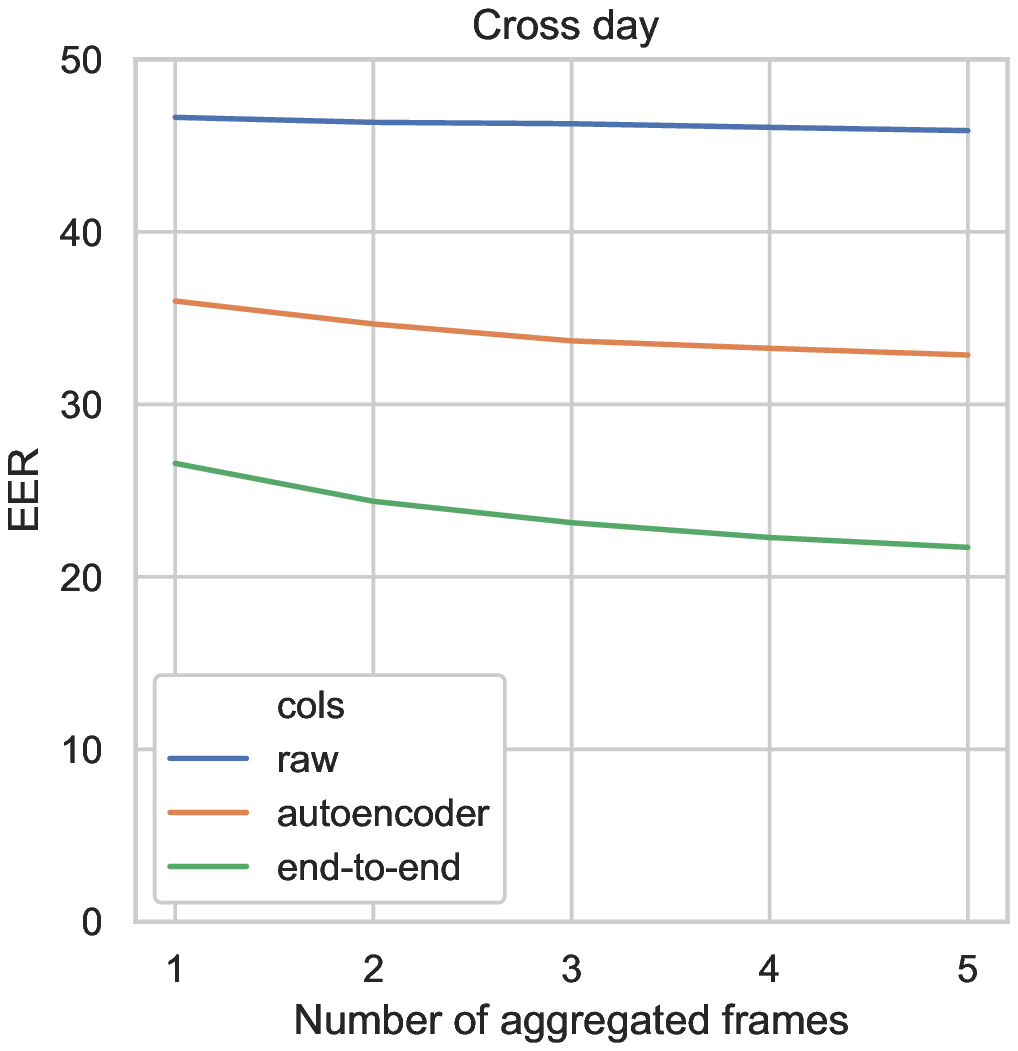}
\caption{Comparison of performances of systems using different features
across different number of aggregated frames. Same-day vs. cross-day
settings.}
\label{fig:aggregation}
\end{subfigure}
\end{figure*}

The systems based on feature learning become more robust with increasing the number of aggregated frame scores. A system that makes decisions based on multiple frames essentially makes the final decision based on more data. Figure \ref{fig:aggregation} shows the performances in terms of EER (the lower, the better) in same-day and cross-day settings, aggregating the scores obtained for 1 to 5 consecutive frames. Interestingly, when raw data is used as features, system performance is unaffected by scores aggregation.

\section{Conclusions}

This study investigated automatic feature learning for a gait verification system. We applied supervised and unsupervised feature learning for gait biometrics and compared their performance metrics. Both types of feature learning models were trained on the same dataset using fully convolutional deep neural network architectures: the supervised setup using an end-to-end model and the unsupervised one using a similarly designed autoencoder. Transfer learning was facilitated by stripping each of the already trained models of several layers to allow using them as universal feature extractors in a gait verification system, therefore exposing them to various sources of gait data. Baseline evaluations show that both type of feature extractors produce robust features, having the supervised setup perform slightly better than the unsupervised one. However, proper data augmentation enabled the autoencoder-based system achieve almost the same performances as the supervised system. The most important conclusion of this work is that autoencoders are a viable alternative for feature learning with the scope of gait verification, having also the advantage of not requiring any labels for training.

\section*{Acknowledgment}
The research has been supported by Sapientia Foundation -- Institute for Scientific Research.


\bibliography{gaitrefs}

\end{document}